\documentclass[runningheads]{llncs}
\usepackage{graphicx}
\usepackage{multirow}
\usepackage{graphics}
\usepackage{amsmath}
\usepackage{multirow}
\usepackage{url}
\usepackage{times}
\usepackage{tabularx}
\usepackage{booktabs,subcaption,amsfonts,dcolumn}
\usepackage{amssymb}
\usepackage{pifont}
\newcommand{\cmark}{\ding{51}}
\newcommand{\xmark}{\ding{55}}
\newcommand{\argmax}{\arg\!\max}
\begin{document}
\title{Techniques for Jointly Extracting Entities and Relations: A Survey}
%
%

\author{Sachin Pawar\inst{1,2} \and Pushpak Bhattacharyya\inst{2} \and Girish K. Palshikar\inst{1}}
\authorrunning{Pawar et al.}
%
\institute{TCS Research \& Innovation, Pune, India-411013 \and Indian Institute of Technology Bombay, Mumbai, India-400076\\
{\tt sachin7.p@tcs.com, pb@cse.iitb.ac.in, gk.palshikar@tcs.com}}
\maketitle              
\begin{abstract}
Relation Extraction is an important task in Information Extraction which deals with identifying semantic relations between entity mentions. Traditionally, relation extraction is carried out after entity extraction in a ``pipeline'' fashion, so that relation extraction only focuses on determining whether any semantic relation exists between a pair of extracted entity mentions. This leads to propagation of errors from entity extraction stage to relation extraction stage. Also, entity extraction is carried out without any knowledge about the relations. Hence, it was observed that jointly performing entity and relation extraction is beneficial for both the tasks. In this paper, we survey various techniques for jointly extracting entities and relations. We categorize techniques based on the approach they adopt for joint extraction, i.e. whether they employ joint inference or joint modelling or both. We further describe some representative techniques for joint inference and joint modelling.  We also describe two standard datasets, evaluation techniques and performance of the joint extraction approaches on these datasets. We present a brief analysis of application of a general domain joint extraction approach to a Biomedical dataset. This survey is useful for researchers as well as practitioners in the field of Information Extraction, by covering a broad landscape of joint extraction techniques.

\keywords{Relation Extraction  \and Entity Extraction \and Joint Modelling \and End-to-end Relation Extraction}
\end{abstract}

\section{Introduction}
Entities such as {\small\sf PERSON} or {\small\sf LOCATION} are the most basic units of information in any natural language text. Mentions of such entities in a sentence are often linked
through well-defined semantic relations (e.g., {\small\sf EMPLOYEE\_OF} relation between a {\small\sf PERSON} and an {\small\sf ORGANIZATION}). The task of Relation Extraction (RE) deals with identifying such relations automatically. Apart from the general domain entities of types such as {\small\sf PERSON} or {\small\sf ORGANIZATION}, there can be domain-specific entities and relations. For example, in  Biomedical domain, an example relation type of interest can be {\small\sf SIDE\_EFFECT} between entities of types {\small\sf DRUG} and {\small\sf ADVERSE\_EVENT}.

A lot of approaches~\cite{zhou2005,jiang2007,bunescu2005,qian2008exploiting,pawar2017relation} have been proposed to address the relation extraction task. Most of these traditional Relation Extraction approaches assume the information about entity mentions is available. Here, information about entity mentions consists of their boundaries (words in a sentence constitute a mention) as well as their entity types. 
Hence, in practice, any \textbf{end-to-end relation extraction} system needs to address 3 sub-tasks: i) identifying boundaries of entity mentions, ii) identifying entity types of these mentions and iii) identifying appropriate semantic relation for each pair of mentions. The first two sub-tasks of end-to-end relation extraction correspond to the {\em Entity Detection and Tracking (EDT)} task defined by the the Automatic Content Extraction (ACE) program~\cite{doddington2004automatic} and the third sub-task corresponds to the {\em Relation Detection and Characterization (RDC)} task. 

Traditionally, the three sub-tasks of end-to-end relation extraction are performed serially in a ``pipeline'' fashion. Hence, the errors in any sub-task are propagated to the subsequent sub-tasks. Moreover, this ``pipeline'' approach only allows {\em unidirectional flow of information}, i.e. the knowledge about entities is used for extracting relations but not vice versa. To overcome these problems, it is necessary to perform some or all of these sub-tasks jointly. In this paper, we survey various end-to-end relation extraction approaches which jointly address entity extraction and relation extraction.

\section{Problem Definition}
The problem of end-to-end relation extraction is defined as follows:

\noindent \textbf{Input}: A natural language sentence $S$

\noindent \textbf{Output}: i) Entity Extraction: List of entity mentions occurring in $S$. Here, each entity mention is identified in terms of its boundaries and entity type. ii) Relation Extraction: List of pairs of entity mentions for which any pre-defined semantic relation holds. 

E.g., $S$ = {\small\tt Paris, John's sister, is staying in New York.} Here, the expected output of an end-to-end relation extraction system is shown in the table~\ref{example_output}. 

\begin{table}\small\center
\caption{Expected output of end-to-end relation extraction system (For definitions of entity and relation types, see Section 7.1)}
\begin{tabular}{ll}
\hline
Entity Extraction & Relation Extraction\\
\hline
{\small\tt Paris} : {\small\sf PER} & $\langle$ {\small\tt Paris}, {\small\tt John} $\rangle$ : {\small\sf PER-SOC}\\
{\small\tt John} : {\small\sf PER} & $\langle$ {\small\tt John}, {\small\tt sister} $\rangle$ : {\sf PER-SOC}\\
{\small\tt sister} : {\small\sf PER} & $\langle$ {\small\tt Paris}, {\small\tt New York} $\rangle$ : {\small\sf PHYS}\\
{\small\tt New York} : {\small\sf GPE} & $\langle$ {\small\tt sister}, {\small\tt New York} $\rangle$ : {\small\sf PHYS}\\
\hline
\end{tabular}
\label{example_output}
\end{table}

\section{Motivating Example}
Any particular semantic relation generally holds between entity mentions of some specific entity types. E.g., social ({\small\sf PER-SOC}) relation holds between two persons ({\small\sf PER}); employee-employer ({\small\sf EMP-ORG}) relation holds between a person ({\small\sf PER}) and an organization ({\small\sf ORG}) or a geo-political entity ({\small\sf GPE}). Hence, information about entity types certainly helps relation extraction. Traditional ``pipeline'' approaches for relation extraction approaches use features based on entity types. However, in these ``pipeline'' approaches there is no bidirectional flow of information; i.e., entity extraction sub-task does not utilize any knowledge / features based on relation information. When entity and relation extraction are jointly addressed, such bidirectional flow is possible. Thus improving performance of both the entity extraction and the relation extraction.

Consider an example sentence: {\small\tt Paris, John's sister, is staying in New York.} Most of the state-of-the-art Named Entity Recognition (NER) tools incorrectly identify {\small\tt Paris} as a mention of type {\small\sf LOC}\footnote{E.g., Stanford CoreNLP 3.9.1 NER identifies {\small\tt Paris} as a city name} and not as {\small\sf PER}. Here, if an entity extraction algorithm has some evidence that {\small\tt Paris} is involved in a social ({\small\sf PER-SOC}) relation, then it would prefer to label {\small\tt Paris} as a {\small\sf PER} than as a {\small\sf LOC}. This is because social relation is only possible between two persons. Thus, the information about relations helps in determining entity types of entity mentions. This is the motivation behind designing algorithms which jointly extract entities and relations.

\section{Overview of Techniques}

\begin{table}[!b]\center\scriptsize
\caption{Overview of various techniques for joint extraction of entities and relations}
\begin{tabular}{lccccccc}
\hline
\multirow{2}{*}{Approach} & Joint & Joint & Model & Inference & \multicolumn{3}{c}{Evaluation}\\
\cline{6-8}
 & Model & Inference & Type & Technique & ACE'04 & ACE'05 & CoNLL'04 \\
\hline
Roth and Yih~\cite{roth2002probabilistic} & \xmark & \cmark &  & Belief Network & \xmark & \xmark & \xmark \\
Roth and Yih~\cite{roth2004linear} & \xmark & \cmark &  & ILP & \xmark & \xmark & \cmark \\
Kate and Mooney~\cite{kate2010joint} & \xmark & \cmark &  & Parsing & \xmark & \xmark & \cmark \\
Chan and Roth~\cite{chan2011exploiting} & \xmark & \cmark &  & Rules & \cmark & \xmark & \xmark \\
Li and Ji~\cite{li2014incremental} & \cmark & \cmark & Structured Prediction & Beam search & \cmark & \cmark & \xmark \\
Miwa and Sasaki~\cite{miwa2014modeling} & \cmark & \cmark & Table+Structured Prediction & Beam search & \xmark & \xmark & \cmark \\
Gupta et al.~\cite{gupta2016table} & \cmark & \xmark & Neural (RNN) &  & \xmark & \xmark & \cmark \\
Pawar et al.~\cite{pawar2016} & \xmark & \cmark &  & MLN & \cmark & \xmark & \xmark \\
Miwa and Bansal~\cite{miwa2016end} & \cmark & \xmark & Neural (Bi/tree-LSTM) &  & \cmark & \cmark & \xmark \\
Pawar et al.~\cite{pawar2017end} & \cmark & \cmark & Table+Neural (NN, LSTM) & MLN & \cmark & \xmark & \xmark \\
Katiyar and Cardie~\cite{katiyar2017going} & \cmark & \xmark & Neural (Bi-LSTM) &  & \cmark & \cmark & \xmark \\
Ren et al.~\cite{ren2017cotype} & \cmark & \xmark & Embedded Representations &  & \xmark & \xmark & \xmark \\
Zheng et al.~\cite{zheng2017joint} & \cmark & \cmark & Neural (Bi-LSTM) & Joint Label & \xmark & \xmark & \xmark \\
Zhang et al.~\cite{zhang2017end} & \cmark & \cmark & Table+Neural (Bi-LSTM) & Global optimization & \xmark & \cmark & \cmark \\
Bekoulis et al.~\cite{bekoulis2018joint} & \cmark & \cmark & CRF, Neural (Bi-LSTM) & Parsing & \cmark & \xmark & \cmark \\
Wang et al.~\cite{wang2018joint} & \cmark & \cmark & Neural (Bi-LSTM) & Parsing & \xmark & \xmark & \xmark \\
Li et al.~\cite{li2017neural} & \cmark & \xmark & Neural (Bi-LSTM) &  & \xmark & \xmark & \xmark \\
\hline
\end{tabular}
\label{tabOverview}
\end{table}

Various techniques have been proposed for jointly extracting entities and relations since 2002. Table~\ref{tabOverview} summarizes most of the techniques from the literature of joint extraction. We visualize each of these techniques from two aspects of joint extraction: {\em joint model} and {\em joint inference}. Most of the techniques exploit either one of these aspects. But some recent techniques have exploited both of these aspects.

Here, by {\em joint model}, we mean that a single model is learned for both the tasks of entity and relation extraction. For example, a single joint neural network model can be learned and both the tasks of entity and relation extraction share the same parameters. Overall, joint models can be of various types as shown in Table~\ref{tabOverview}. Moreover, by {\em joint inference}, we mean that the decision about entity and relation labels is taken jointly at a global level (usually a sentence). Here, there may be separate underlying local models for entity and relation extraction. Overall, there are several joint inference / decoding techniques which are shown in Table~\ref{tabOverview}.

\section{Joint Inference Techniques}
Here, we describe a few techniques used for joint inference:

\noindent \textbf{Integer Linear Programming (ILP):} Here, a {\em global} decision is taken by using Integer Linear Programming which is consistent with some domain constraints . This approach was proposed by Roth and Yih~\cite{roth2004linear}. They first learn independent {\em local} classifiers for entity and relation extraction. During inference, given a sentence, a global decision is produced such that the domain-specific or task-specific constraints are satisfied. Often these constraints capture mutual compatibility of entity and relation types. A simple example of such constraints is: both the arguments of the {\small\sf PER-SOC} relation should be {\small\sf PER}. Consider our example sentence -- {\small\tt Paris, John's sister, is staying in New York.} Here, the entity extractor identifies two mentions {\small\tt John} and {\small\tt Paris} and also predicts entity types for these mentions. For {\small\tt John}, let the predicted probabilities be : Pr({\small\sf PER}) = 0.99 and Pr({\small\sf ORG}) = 0.01. For {\small\tt Paris}, let the predicted probabilities be : Pr({\small\sf GPE}) = 0.75 and Pr({\small\sf PER}) = 0.25. Also, the relation extractor identifies the relation {\small\sf PER-SOC} between the two mentions. If we accept the best suggestions given by the local classifiers, then the global prediction is that the relation {\small\sf PER-SOC} exists between the {\small\sf PER} mention {\small\tt John} and the {\small\sf GPE} mention {\small\tt Paris}. But this violates the domain-constraint mentioned earlier. Hence the global decision which satisfies all the specified constraints would be to label both the mentions as {\small\sf PER} and mark the {\small\sf PER-SOC} relation between them.

\noindent \textbf{Markov Logic Networks (MLN):} Similar to ILP, MLN provides another framework for taking a global decision consistent with the domain constraints. MLN combines first order logic with probability. The domain rules or domain knowledge is represented in an MLN using weighted first order logic rules. Stronger the belief about any rule, higher is its associated weight. Inference in such an MLN gives the most probable true groundings of certain (query) predicates, while ensuring maximum weighted satisfiability of the rules. Pawar et al.~\cite{pawar2016,pawar2017end}, use MLN for joint inference for extracting entities and relations. As compared to ILP, MLN provides better representability in the form of first order logic rules. For example, the above-mentioned rule ``both the arguments of the {\small\sf PER-SOC} relation should be {\small\sf PER}'' can be written as:
\begin{center}
{\small\sf PER-SOC}$(x,y) \Rightarrow$ {\small\sf PER}$(x)\wedge$ {\small\sf PER}$(y)$
\end{center}
In addition to the rules for ensuring compatibility of entity and relation types, MLN can easily represent other complex domain knowledge. For example, the rule ``a person can be employed at only one organization at a time'' can be written as:
\begin{center}
{\small\sf EMP-ORG}$(x,y)\wedge$ {\small\sf PER}$(x)\wedge$ {\small\sf ORG}$(y)\wedge$ {\small\sf ORG}$(z)\wedge (y\neq z)  \Rightarrow \neg${\small\sf EMP-ORG}$(x,z)$
\end{center}

\noindent \textbf{Joint Label:} Zheng et al.~\cite{zheng2017joint} proposed a novel tagging scheme for joint extraction of entities and relations. This tagging scheme reduces the joint extraction task to a tagging problem. Intuitively, a single tag is assigned to a word which encodes entity as well as relation label information, automatically leading to joint inference. 
\begin{figure}
\includegraphics[width=\linewidth,height=2cm]{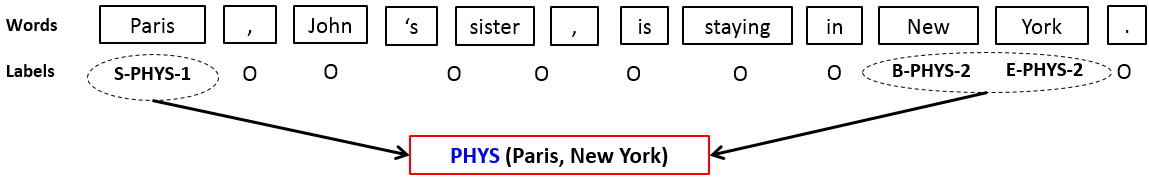}
\caption{\label{figTaggingScheme} The proposed new tagging scheme for an relation instance {\small\sf PHYS} ({\small\tt Paris}, {\small\tt New York}) in our example sentence.}
\end{figure}
Figure~\ref{figTaggingScheme} depicts an example sentence and its annotations as per the proposed new tagging scheme. The tag ``O'' represents the ``Other'' tag, which means that the corresponding word is not part of the expected relation tuples. The other tags consist of three parts: i) the word position in the entity, ii) the relation type, and iii) the relation role (argument number). The \textbf{BIES} (\textbf{B}egin, \textbf{I}nside, \textbf{E}nd, \textbf{S}ingle) encoding scheme is used for marking entity boundaries. The relation type information is obtained from a predefined set of relations and the relation role information is represented by the numbers 1 and 2. 
Let $Entity_1$ and $Entity_2$ be the first and second entity arguments of a relation type $RT$, respectively. Words in $Entity_1$ are marked with the relation role 1 for $RT$. Similarly, words in $Entity_2$ are marked with the relation role 2. Hence, the total number of tags is $2\times 4\times NumRelationTypes + 1$. Here, the multiplier $4$ represents the entity boundary tags \textbf{BIES} and other multiplier $2$ represents two entity arguments for each relation type. For example shown in Figure~\ref{figTaggingScheme}, the words {\small\tt New} and {\small\tt York} are part of the entity mention which is the second argument of {\small\sf PHYS} relation and hence are marked with the tags {\small\sf B-PHYS-2} and {\small\sf E-PHYS-2}, respectively. One limitation of this approach is that currently it can not model the scenario where a single entity mention is involved in multiple relations with multiple other entity mentions. Hence, the other relation {\small\sf PER-SOC} ({\small\tt Paris}, {\small\tt John}) in which {\small\tt Paris} is involved, can not be handled.

\noindent \textbf{Beam Search:}
Li and Ji~\cite{li2014incremental} proposed an approach for {\em incremental} joint extraction of entities and relations. They formulated the joint extraction task as a structured prediction problem to reveal the linguistic and logical properties of the hidden structures. Here, the output structure of each sentence was interpreted as a graph in which entity mentions are nodes and relations are directed arcs labelled with relation types. They designed several local as well as global features to characterize and score these structures. Hence, the joint extraction problem was reduced to predicting a structure with the highest score for any given sentence. They proposed a joint decoding / inference approach for this structured prediction task using beam-search. Intuitively, at the $i^{th}$ token, $k$ best partial assignments / structures are maintained and extended further. Similarly, beam-search based inference was also employed by Miwa and Sasaki~\cite{miwa2014modeling} where the output structure for a sentence was a table representation.

\begin{figure}[t]\center
\includegraphics[scale=0.3]{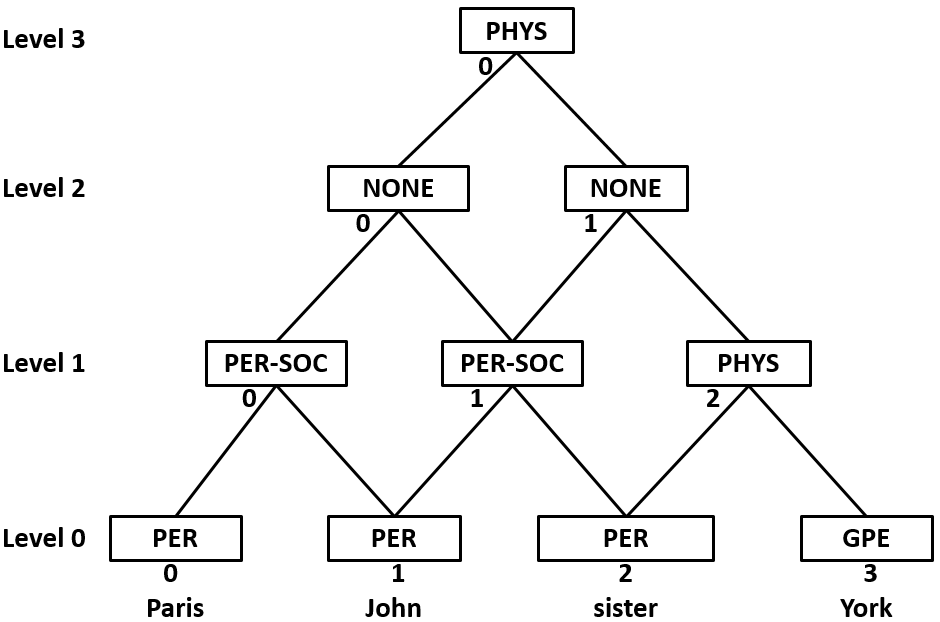}
\caption{Card-pyramid graph for the sentence {\small\tt Paris, John's sister, is staying in New York.}}
\label{fig_card_pyramid}
\end{figure}

\noindent \textbf{Parsing:} Kate and Mooney~\cite{kate2010joint} proposed a parsing based approach which uses a graph called as {\em card-pyramid}. The graph is so called because it encodes mutual dependencies among the entities and relations in a graph structure which resembles pyramid constructed using playing cards. This is a tree-like graph which has one root at the highest level, internal nodes at intermediate levels and leaves at the lowest level. Each entity in the sentence correspond to one leaf and if there are $n$ such leaves then the graph has $n$ levels. Each level $l$ contains one less node than the number of nodes in the $(l-1)$ level. The node at position $i$ in level $l$ is parent of nodes at positions $i$ and $(i+1)$ in the level $(l-1)$. Each node in the higher layers (i.e. layers except the lowest layer), corresponds to a possible relation between the leftmost and rightmost nodes under it in the lowest layer. Figure~\ref{fig_card_pyramid} shows this {\em card-pyramid} graph for an example sentence. 
To jointly label the nodes in the card-pyramid graph, the authors propose a parsing algorithm analogous to the bottom-up CYK parsing algorithm for Context Free Grammar (CFG) parsing. The grammar required for this new parsing algorithm is called Card-pyramid grammar and its consists of following production types:
\begin{itemize}
\item Entity Productions: These are of the form $EntityType\rightarrow Entity$, e.g. {\small\sf PER}$\rightarrow${\small\tt John}. Similar to the ILP based approach, a local entity classifier is trained to compute the probability that entity in the RHS being of the type given in the LHS of the production.
\item Relation Productions: These are of the form $RelationType\rightarrow EntityType1$ $EntityType2$, e.g. {\small\sf PHYS}$\rightarrow${\small\sf PER} {\small\sf GPE}. A local relation classifier is trained to obtain the probability that the two entities in the RHS are related by the type given in the LHS of the production.
\end{itemize}
Given the entities in a sentence, the Card-pyramid grammar, and the local entity and relation classifiers, the card-pyramid parsing algorithm attempts to find the most probable labelling of all of its nodes which corresponds the entity and relation types. One limitation of this approach is that only entity type identification happens jointly with relation classification, i.e. boundary detection of entity mentions should be done as a pre-processing step and does not happen jointly. Recently, Bekoulis et al.~\cite{bekoulis2018joint} and Wang et al.~\cite{wang2018joint} proposed joint extraction techniques which use dependency parsing like approaches for joint inference. Also, they allow multiple heads for a node (word) to represent participation in multiple relations simultaneously with other nodes.

\section{Joint Models}
Here, we describe a few joint models which have been employed for joint extraction of entities and relations:

\noindent \textbf{Structured Prediction:} In most of the earlier approaches for joint extraction of entities and relations, it was assumed that the boundaries of the entity mentions are known. Li and Ji~\cite{li2014incremental} presented an incremental joint framework for simultaneous extraction of entity mentions and relations, which also incorporates the problem of boundary detection for entity mentions. The authors proposed to formulate the problem of joint extraction of entities and relations as a structured prediction problem. They aimed to predict the output structure ($y \in Y$) for a given sentence ($x\in X$), where this structure can be viewed as a graph modelling entity mentions as nodes and relations as directed arcs with relation types as labels. Following linear model is used to predict the most probable structure $y'$ for $x$ where $f(x,y)$ is the feature vector that characterizes the entire structure.
\begin{equation*}
y' = \argmax_{y \in Y(x)} f(x,y)\cdot W
\end{equation*}
The score of each candidate assignment is defined as the inner product of the feature vector $f(x,y)$ and feature weights $W$. The number of all possible structures for any given sentence can be very large and there does not exist a polynomial-time algorithm to find the best structure. Hence, they apply beam-search to expand partial configurations for the input sentence incrementally to find the structure with the highest score.

\begin{figure}[t]
\includegraphics[width=\linewidth,height=0.45\linewidth]{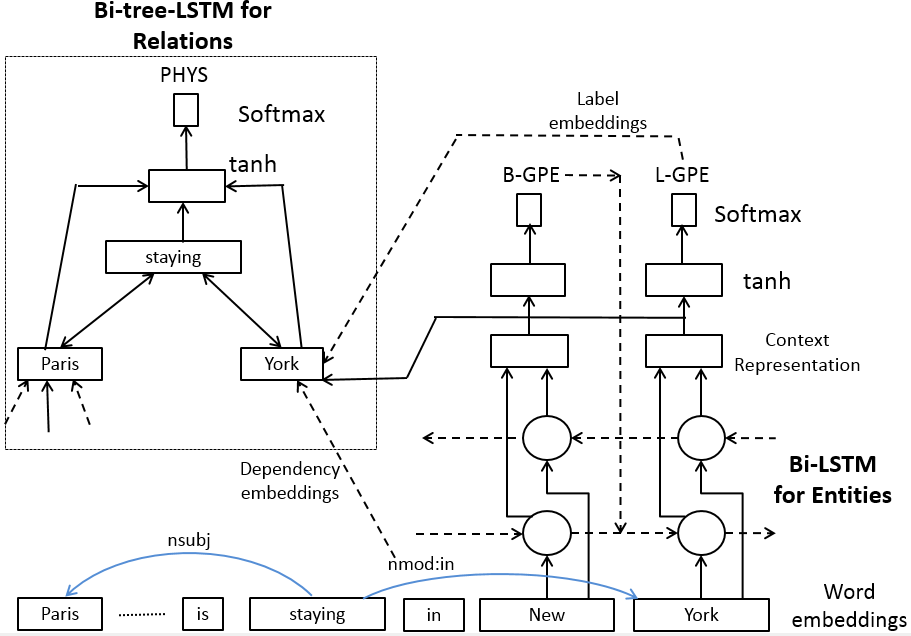}
\caption{End-to-end relation extraction model, with bidirectional sequential and bidirectional tree-structured LSTM-RNNs.}
\label{figMiwaModel}
\end{figure}

\noindent \textbf{Neural Models:} Here, predictions for both the tasks of entity and relation extraction are carried out using a single joint neural model, where at least some of the model parameters are shared across both the tasks. Joint modelling is realized through such parameter sharing where training for any task updates the parameters involved in both the tasks. Miwa and Bansal~\cite{miwa2016end} presented a neural model for capturing both word sequence and dependency tree substructure information by stacking bidirectional tree-structured LSTMs (tree-LSTM) on bidirectional sequential LSTMs (Bi-LSTM). Their model jointly represents both entities and relations with shared parameters in a single model. The overview of the model is illustrated in the Figure~\ref{figMiwaModel}. It consists of three representation layers: i) a word embeddings layer, ii) a word sequence based LSTM-RNN layer (sequence layer), and iii) a dependency subtree based LSTM-RNN layer (dependency layer). While decoding, entities are detected in greedy, left-to-right manner on the sequence layer. And relation classification is carried out on the dependency layers, where each subtree based LSTM-RNN corresponds to a relation candidate between two detected entities. After decoding the entire model structure, the parameters are updated simultaneously via backpropagation through time (BPTT). The dependency layers are stacked on the sequence layer, so the embedding and sequence layers are shared by both entity detection and relation classification, and the shared parameters are affected by both entity and relation labels. This is the first joint neural model which motivated several other joint models~\cite{katiyar2017going,zheng2017joint}. This model was adopted for Biomedical domain by Li et al.~\cite{li2017neural}. In addition, they use Convolutional Neural Networks (CNN) for extracting morphological information (like prefix or suffix) from characters of words. Then each word in a sentence is represented by a concatenated vector of its word embeddings, POS embeddings and character-level representation by CNN. Character-level information is more useful in Biomedical domain because several biological entities share morphological or orthographic features, e.g., bacteria names {\small\tt helicobacter} and {\small\tt campylobacter} share the suffix {\small\tt bacter}.

\noindent \textbf{Table Representation:} Another idea for jointly modelling entity and relation extraction tasks is {\em Table Representation} or {\em Table Filling}. It was first proposed by Miwa and Sasaki~\cite{miwa2014modeling}. Here, a table is associated with each sentence where every table cell is labelled with an appropriate label so that the whole entities and relations structure in a sentence is represented in a single table. Table~\ref{tableRepresentation} depicts this table representation for an example sentence. The diagonal cells of the table represent the entity labels which capture both boundary and type information with the help of BILOU (\textbf{B}egin, \textbf{I}nside, \textbf{L}ast, \textbf{O}utside, \textbf{U}nit) or BIO encoding. E.g., in Table~\ref{tableRepresentation}, the word {\small\tt New} gets the label {\small\tt B-GPE} as it is the first word of the complete entity mention {\small\tt New York}. The off-diagonal cells represent relation labels. Here, relations between entity mentions are mapped to relations between the last words of the mentions. E.g., the {\small\sf PHYS} relation between {\small\tt sister} and {\small\tt New York} is assigned to the cell corresponding to {\small\tt sister} and {\small\tt York}. $\perp$ represents no pre-defined relation between the corresponding words. As the table is symmetric, only upper or lower triangular part of the table needs to be labelled. Miwa and Sasaki~\cite{miwa2014modeling} approach this table filling problem using structure learning approach similar to Li and Ji~\cite{li2014incremental}. They define a scoring function to evaluate a possible label assignment to a table and build a model which predicts the most probable label assignment for a table which maximizes the scoring function. During inference, beam search is used which assigns labels to cells one by one and keeps the top $K$ best assignments when moving from a cell to the next cell. Finally, it returns the best assignment when labels are assigned to all the cells. The authors propose various strategies to arrange the cells in two dimensions to a linear order. They also integrate various label dependencies into the scoring function to avoid illegal label assignments. E.g., cell corresponding to the $i^{th}$ and $j^{th}$ words should never be assigned any valid relation label if any of the words are labelled with entity label {\small\sf O}.
\begin{table}[t]\small\center
\caption{Table Representation for an example sentence}
\begin{tabular}{|c|c|c|c|c|c|c|c|c|c|c|c|c|}
\hline
 & {\tt Paris} & {\tt ,} & {\tt John} & {\tt 's} & {\tt sister} & {\tt ,} & {\tt is} & {\tt staying} & {\tt in} & {\tt New} & {\tt York} & {\tt .} \\
\hline
{\tt Paris} & {\sf U-PER} & $\perp$ & {\sf PER-SOC} & $\perp$ & $\perp$ & $\perp$ & $\perp$ & $\perp$ & $\perp$ & $\perp$ & {\sf PHYS} & $\perp$\\
\hline
{\tt ,} &  & {\sf O} & $\perp$ & $\perp$ & $\perp$ & $\perp$ & $\perp$ & $\perp$ & $\perp$ & $\perp$ & $\perp$ & $\perp$\\
\hline
{\tt John} &  &  & {\sf U-PER} & $\perp$ & {\sf PER-SOC} & $\perp$ & $\perp$ & $\perp$ & $\perp$ & $\perp$ & $\perp$ & $\perp$\\
\hline
{\tt 's} &  &  &  & {\sf O} & $\perp$ & $\perp$ & $\perp$ & $\perp$ & $\perp$ & $\perp$ & $\perp$ & $\perp$\\
\hline
{\tt sister} &  &  &  &  & {\sf U-PER} & $\perp$ & $\perp$ & $\perp$ & $\perp$ & $\perp$ & {\sf PHYS} & $\perp$\\
\hline
{\tt ,} &  &  &  &  &  & {\sf O} & $\perp$ & $\perp$ & $\perp$ & $\perp$ & $\perp$ & $\perp$\\
\hline
{\tt is} &  &  &  &  &  &  & {\sf O} & $\perp$ & $\perp$ & $\perp$ & $\perp$ & $\perp$\\
\hline
{\tt staying} &  &  &  &  &  &  &  & {\sf O} & $\perp$ & $\perp$ & $\perp$ & $\perp$\\
\hline
{\tt in} &  &  &  &  &  &  &  &  & {\sf O} & $\perp$ & $\perp$ & $\perp$\\
\hline
{\tt New} &  &  &  &  &  &  &  &  &  & {\sf B-GPE} & $\perp$ & $\perp$\\
\hline
{\tt York} &  &  &  &  &  &  &  &  &  &  & {\sf L-GPE} & $\perp$\\
\hline
{\tt .} &  &  &  &  &  &  &  &  &  &  &  & {\sf O}\\
\hline
\end{tabular}
\label{tableRepresentation}
\end{table}

The table representation idea has further motivated several other joint extraction approaches. Pawar et al.~\cite{pawar2017end} use a similar table representation but instead of using BILOU encoding to represent entity boundaries, they introduced a new relation {\small\sf WEM} (Within Entity Mention) between head word\footnote{Head word is generally the last word of noun phrase entities but not always. E.g., for {\tt Bank of America}, the head word is {\tt Bank}. Head word is that word of an entity mention through which the mention is linked to the rest of the sentence in its dependency tree.} of an entity mention and other words in the same entity mention. E.g., they would assign entity labels {\small\sf O} and {\small\sf GPE} to the words {\small\tt New} and {\small\tt York}, respectively and assign relation label {\small\tt WEM} to the cell corresponding to {\small\tt New} and {\small\tt York}. Further, they train a neural network based model to predict an appropriate label for each cell in the table. They also employ Markov Logic Networks (MLN) based inference at a sentence level to incorporate various dependencies among entity and relation labels. Other recent approaches proposed by Zhang et al.~\cite{zhang2017end} and Gupta et al.~\cite{gupta2016table} build upon the same table representation idea and use Recursive Neural Networks (RNN) and Long Short-Term Memory (LSTM) based models.

\section{Experimental Evaluation}
In this section, we describe some of the most widely used datasets for end-to-end relation extraction and summarize the reported results on those datasets. We also describe the evaluation methodology and other experimental analysis.
\subsection{Datasets}
\subsubsection{ACE 2004:} It is the most widely used dataset in the relation extraction literature and is available from Linguistic Data Consortium (LDC) as catalogue LDC2005T09. It annotates both entity and relation types information in an XML like format. It identifies 7 entity types\footnote{\url{www.ldc.upenn.edu/sites/www.ldc.upenn.edu/files/english-edt-v4.2.6.pdf}}: (i) {\small\sf PER} (person), (ii) {\small\sf ORG} (organization), (iii) {\small\sf LOC} (location), (iv) {\small\sf GPE} (geo-political entity), (v) {\small\sf FAC} (facility), (vi) {\small\sf VEH} (vehicle), and (vii) {\small\sf WEA} (weapon). Additionally, it identifies 22 fine-grained relation types which are grouped into 6 coarse-grained relation types\footnote{\url{www.ldc.upenn.edu/sites/www.ldc.upenn.edu/files/english-rdc-v4.3.2.PDF}}: (i) {\small\sf EMP-ORG} (employee-organization or subsidiary relationships), (ii) {\small\sf GPE-AFF} (affiliations of {\small\sf PER/ORG} to an {\small\sf GPE} entity), (iii) {\small\sf PER-SOC} (social relationships between two {\small\sf PER} entities), (iV) {\small\sf ART} (agent-artifact relationship), (v) {\small\sf PHYS} (physical / located at), (vi) {\small\sf OTHER-AFF} (other {\small\sf PER/ORG} affiliations). Chan and Roth~\cite{chan2011exploiting} used this dataset for the first time for evaluating end-to-end relation extraction. They ignored the original {\small\sf DISC} (discourse) relation as it was only for the purpose of the discourse. They used only news wire and broadcast news subsections of this dataset which consists of 345 documents and 4011 positive relation instances. All the later approaches followed the same methodology for producing comparable results.
\subsubsection{ACE 2005:} This dataset~\cite{walker2006ace} is also available from LDC as catalogue LDC2006T06. It annotates the same entity types a that of ACE 2004. ACE 2005 also kept the relation types {\small\sf PER-SOC}, {\small\sf ART} and {\small\sf GPE-AFF} of ACE 2004, but it split {\small\sf PHYS} into two relation types {\small\sf PHYS} and a new relation type {\small\sf PART-WHOLE}. The {\small\sf DISC} relation type was removed, and the relation type {\small\sf OTHER-AFF} was merged into {\small\sf EMP-ORG}. It was observed that ACE 2005 improved on both annotation quality and relation type definition, as compared to ACE 2004. Li and Ji~\cite{li2014incremental} used this dataset for the first time for evaluating end-to-end relation extraction. Ignoring two small subsets ({\em cts} and {\em un}) from informal genres, they selected the remaining 511 documents. These were randomly split into 3 parts: (i) training (351), (ii) development (80), and (iii) blind test set (80). All the later approaches followed the same methodology for producing comparable results.

\subsection{Evaluation of End-to-End Relation Extraction}
As discussed earlier, the end-to-end relation extraction system is expected to identify:  (i) boundaries of entity mentions, (ii) entity types of these entity mentions, and (iii) relation type (if any) for each pair of entity mentions. Hence, evaluation of end-to-end relation extraction is often done at 2 levels:
\begin{enumerate}
\item \textbf{Entity extraction:} Here, only entity extraction performance is evaluated. Two entity mentions are said to be {\em matching} if both have same boundaries (i.e. contain exactly the same sequence of words) and same entity type. Any predicted entity mention is counted as a true positive (TP) if it matches with any of the gold-standard entity mentions in the same sentence, otherwise it is counted as a false positive (FP). Both TP or FP are counted for the predicted entity type. Similarly, for each gold-standard entity mention, if there no matching predicted entity mention in the same sentence, then a false negative (FN) is counted for the gold-standard entity type. For each entity type, precision, recall and F1 are computed using its TP, FP and FN counts. F1-scores across all entity types are micro-averaged for computing overall entity extraction performance.
\item \textbf{Entity+Relation extraction:} Here, end-to-end relation extraction performance is evaluated. Any predicted or gold-standard relation mention consists of a pair of entity mentions along with their entity types, and an associated relation type. Hence, two relation mentions are said to be {\em matching} only if both the entity mentions match and associated relation types are same. Each gold-standard relation mention is counted as a TP if there is a {\em matching} predicted relation mention, otherwise it is counted as FN. Similarly, each predicted relation mention is counted as an FP unless there is any {\em matching} gold-standard relation mention. For each relation type, precision, recall and F1 are computed using its TP, FP and FN counts. F1-scores across all relation types are micro-averaged for computing overall entity extraction performance.
\end{enumerate}

\begin{table}[t]\center
\caption{Performance of various approaches on the ACE 2004 dataset. The numbers are micro-averaged and obtained after 5-fold cross-validation. Actual folds used by each approach may differ.}
\begin{tabular}{lccccccc}
\hline
{\bf Approach} & \multicolumn{3}{c}{\bf Entity Extraction} & \hspace{2mm} & \multicolumn{3}{c}{\bf Entity+Relation Extraction}\\
\cline{2-4}\cline{6-8}
  & \hspace{4mm}{\bf P}\hspace{4mm} & {\bf R} & {\bf F} &  & \hspace{6mm}{\bf P}\hspace{4mm} & {\bf R} & {\bf F}\\
\hline
Pipeline~\cite{li2014incremental} & 81.5 & 74.1 & 77.6 &  & 58.4 & 33.9 & 42.9 \\
Chan and Roth~\cite{chan2011exploiting} &  &  &  &  & 42.9 & 38.9 & 40.8\\
Li and Ji~\cite{li2014incremental} & 83.5 & 76.2 & 79.7 &  & 60.8 & 36.1 & 45.3 \\
Pawar et al.~\cite{pawar2016} & 79.0 & 80.1 & 79.5 &  & 52.4 & 41.3 & 46.2 \\
Miwa and Bansal~\cite{miwa2016end} & 80.8 & 82.9 & {\bf81.8} &  & 48.7 & 48.1 & 48.4 \\
Pawar et al.~\cite{pawar2017end} & 81.2 & 79.7 & 80.5 &  & 56.7 & 44.5 & {\bf 49.9} \\
Katiyar and Cardie~\cite{katiyar2017going} & 81.2 & 78.1 & 79.6 &  & 46.4 & 45.3 & 45.7 \\
Bekoulis et al.~\cite{bekoulis2018joint} & 81.0 & 81.3 & 81.2 &  & 50.1 & 44.5 & 47.1 \\
\hline
\end{tabular}
\label{resultsACE2004}
\end{table}

\begin{table}[!b]\center
\caption{Performance of various approaches on the ACE 2005 dataset. The numbers are micro-averaged and obtained on a test split of 80 documents. The (\_) performance numbers are not reported in the original paper.}
\begin{tabular}{lccccccc}
\hline
{\bf Approach} & \multicolumn{3}{c}{\bf Entity Extraction} & \hspace{2mm} & \multicolumn{3}{c}{\bf Entity+Relation Extraction}\\
\cline{2-4}\cline{6-8}
  & \hspace{4mm}{\bf P}\hspace{4mm} & {\bf R} & {\bf F} &  & \hspace{6mm}{\bf P}\hspace{4mm} & {\bf R} & {\bf F}\\
\hline
Pipeline~\cite{li2014incremental} & 83.2 & 73.6 & 78.1 &  & 65.1 & 38.1 & 48.0 \\
Li and Ji~\cite{li2014incremental} & 85.2 & 76.9 & 80.8 &  & 65.4 & 39.8 & 49.5 \\
Miwa and Bansal~\cite{miwa2016end} & 82.9 & 83.9 & 83.4 &  & 57.2 & 54.0 & 55.6 \\
Katiyar and Cardie~\cite{katiyar2017going} & 84.0 & 81.3 & 82.6 &  & 55.5 & 51.8 & 53.6 \\
Zhang et al.~\cite{zhang2017end} & \_ & \_ & \textbf{83.6} &  & \_ & \_ & \textbf{57.5} \\
\hline
\end{tabular}
\label{resultsACE2005}
\end{table}

\subsubsection{Analysis of Results:} Tables~\ref{resultsACE2004} and~\ref{resultsACE2005} show the results of various approaches on the ACE 2004 and ACE 2005 datasets, respectively. The F1-scores still below 60\% indicate how challenging the task of end-to-end relation extraction is. Li and Ji~\cite{li2014incremental} carried out an interesting experiment where two human annotators were asked to perform end-to-end relation extraction manually on the ACE 2005 test dataset. The human F1-score for this task was observed to be around 70\%. Moreover, F1-score of the inter-annotator agreement (the entity / relation extractions where both the annotators agreed) was only about 51.9\%. This analysis clearly establishes the high difficulty level of the task.

\begin{table}[t]\small\center
\caption{Example sentences from the ACE 2004 dataset illustrating how the joint extraction of entities and relations helps in determining entity type of {\small\tt \textbf{its}}. Entity mentions of interest are highlighted in \textbf{bold}}
\begin{tabular}{p{0.05\linewidth}p{0.95\linewidth}}
\hline
\multirow{2}{*}{S1} & {\small\tt U.S. District Court Judge Murray Schwartz in Wilmington, Del., ruled that Camelot Music could not deduct interest on loans it took out against life insurance on \textbf{its} 1,430 \textbf{employees} in 1990 through 1993.} \\
\cline{2-2}
 & EntityType ({\small\tt its}) = {\small\sf ORG}, EntityType ({\small\tt employees}) = {\small\sf PER},\newline RelType ($\langle${\small\tt its}, {\small\tt employees}$\rangle$) = {\small\sf EMP-ORG} \\
\hline
\multirow{2}{*}{S2} & {\small\tt our choice is the choice of permanent, comprehensive and just peace, and our aim is to liberate our land and to create our independence state in palestinian blast land with jerusalem as \textbf{its} \textbf{capital} and the return of our refugees to their homes.} \\
\cline{2-2}
 & EntityType ({\small\tt its}) = {\small\sf GPE}, EntityType ({\small\tt capital}) = {\small\sf GPE},\newline RelType ($\langle${\small\tt its}, {\small\tt capital}$\rangle$) = {\small\sf PHYS}\\
\hline
\end{tabular}
\label{exSentences}
\end{table}
Another important aspect of the ACE datasets to note is the nature of its entity mentions. Overall, three types of entity mentions are annotated in the ACE datasets: (i) name mentions (generally proper nouns, e.g. {\small\tt John}, {\small\tt United States}), (ii) nominal mentions (generally common nouns, e.g. {\small\tt guy}, {\small\tt employee}), and (iii) pronoun mentions (e.g. {\small\tt he}, {\small\tt they}, {\small\tt it}). Unlike the traditional Named Entity Recognition (NER) task which extracts only the name mentions, the ACE entity extraction task focusses on extracting all the three types of mentions. This makes it more challenging task yielding lower accuracies. Especially for pronoun mentions like {\small\tt its} in the example sentences in Table~\ref{exSentences}, determining the entity type is more challenging. This is because, the mention {\small\tt its} is observed both as {\small\sf ORG} or as {\small\sf GPE} in the training data depending on the context. In the sentence S1 in Table~\ref{exSentences}, the knowledge that {\small\tt its} is related to {\small\tt employees} through the {\small\sf EMP-ORG} relation, helps in labelling {\small\tt its} as {\small\sf ORG}. Similarly, in the sentence S2, the knowledge that {\small\tt its} is related to {\small\tt capital} through the {\small\sf PHYS} relation, helps in labelling {\small\tt its} as {\small\sf GPE}. Hence, these examples illustrate that unlike pipeline methods, in joint extraction methods, both the tasks of entity extraction and relation extraction help each other.

\subsection{Domain-specific Entities and Relations}
Except Li et al.~\cite{li2017neural}, all other joint extraction approaches in Table~\ref{tabOverview} are evaluated on {\em general} domain datasets like ACE 2004 or ACE 2005. There is no previous study on how well the approaches designed for general domain work for domain-specific entities and relations. In this section, we present the results of our experiments where we apply a general domain technique on a Biomedical dataset. As a representative general domain approach, we choose Pawar et al.~\cite{pawar2017end} which is the best performing approach on the ACE 2004 dataset. 

\begin{table}[!t]\center
\caption{Performance of various approaches on the ADE dataset. The numbers are micro-averaged and obtained using 10-fold cross-validation. Actual folds used by each approach may differ.}
\begin{tabular}{lccccccc}
\hline
{\bf Approach} & \multicolumn{3}{c}{\bf Entity Extraction} & \hspace{2mm} & \multicolumn{3}{c}{\bf Entity+Relation Extraction}\\
\cline{2-4}\cline{6-8}
  & \hspace{4mm}{\bf P}\hspace{4mm} & {\bf R} & {\bf F} &  & \hspace{6mm}{\bf P}\hspace{4mm} & {\bf R} & {\bf F}\\
\hline
Li et al.~\cite{li2017neural} & 82.7 & 86.7 & 84.6 &  & 67.5 & 75.8 & \textbf{71.4} \\
Pawar et al.~\cite{pawar2017end} (GloVe vectors) & 80.0 & 82.4 & 81.2 &  & 65.8 & 66.6 & 66.2 \\
Pawar et al.~\cite{pawar2017end} (PubMed vectors) & 82.1 & 84.0 & 83.0 &  & 68.5 & 68.0 & 68.2 \\
Pawar et al.~\cite{pawar2017end} (GloVe vectors, Lenient) & 82.8 & 85.2 & 84.0 &  & 70.6 & 71.3 & 70.9 \\
Pawar et al.~\cite{pawar2017end} (PubMed vectors, Lenient) & 85.0 & 86.8 & 85.9 &  & 73.0 & 73.7 & 73.3 \\
\hline
\end{tabular}
\label{resultsADE}
\end{table}

Li et al.~\cite{li2017neural} evaluate their end-to-end relation extraction approach on the \textbf{Adverse Drug Event (ADE)} dataset~\cite{gurulingappa2012development}. This dataset contains sentences from PubMed abstracts annotated with entity types {\small\sf DRUG}, {\small\sf ADVERSE\_EVENT} and {\small\sf DOSAGE}. It also contains annotations for two relation types: (i) {\small\sf DRUG-AE} between a {\small\sf DRUG} and an {\small\sf ADVERSE\_EVENT} it causes, and (ii) {\small\sf DRUG-DOSAGE} between a {\small\sf DRUG} and its {\small\sf DOSAGE}. Li et al.~\cite{li2017neural} evaluated their model only on a subset of the ADE dataset containing sentences with at least one instance of the {\small\sf DRUG-AE} relation. They also ignored 120 relation instances containing nested gold annotations, e.g., {\small\tt lithium intoxication}, where {\small\tt lithium} causes {\small\tt lithium intoxication}. We also followed the same methodology for creating a dataset for our experiments. We ended up with a dataset of 4228 distinct sentences\footnote{Li et al.~\cite{li2017neural} mentions number of sentences in their dataset to be 6821 which seems to be a typo because the original paper~\cite{gurulingappa2012development} for ADE dataset mentions that there are only 4272 sentences containing at least one drug-related adverse effect mention. After ignoring the 120 relation instances of nested annotations, this number comes down to 4228 in our dataset.} containing 6714 relation instances. Following is an example sentence and annotations from this dataset: {\small\tt After infliximab treatment, additional sleep studies revealed an increase in the number of apneic events and SaO2 dips suggesting that TNFalpha plays an important role in the pathophysiology of sleep apnea.} There are two annotated relation instances of {\small\sf DRUG-AE} for this sentence: (i) $\langle${\small\tt infliximab}, {\small\tt increase in the number of apneic events}$\rangle$, and (ii) $\langle${\small\tt infliximab}, {\small\tt SaO2 dips}$\rangle$.

\subsubsection{Analysis of Results:} Table~\ref{resultsADE} shows the results of both the methods (Li et al.~\cite{li2017neural} and Pawar et al.~\cite{pawar2017end}) on the ADE dataset for end-to-end extraction of the {\small\tt DRUG-AE} relation. Li et al. used 300 dim word embeddings pre-trained on PubMed corpus~\cite{pyysalodistributional}. For Pawar et al., we experimented with two types of word embeddings: 100-dim GloVe embeddings trained on Wikipedia corpus~\cite{pennington2014glove} (as reported in the original paper) as well as 300-dim embeddings trained on PubMed corpus~\cite{pyysalodistributional}. As the ADE dataset is also derived from PubMed abstracts, the PubMed word embeddings perform better than GloVe embeddings. Even though it is designed for the general domain, Pawar et al.~\cite{pawar2017end} produces comparable results with respect to Li et al.~\cite{li2017neural}. Upon detailed analysis of errors, we found that the major source of errors was incorrect boundary detection for entities of type {\small\sf ADVERSE\_EVENT}. As compared to ACE datasets, the entities in the ADE dataset can have more complex syntactic structures. E.g., it is very rare in case of the ACE entities to be noun phrases (NP) subsuming prepositional phrases (PP), but in the ADE dataset, we frequently encounter entities like {\small\tt increase in the number of apneic events}. We also observed that the boundary annotations for the {\small\sf ADVERSE\_EVENT} entities are inconsistent. E.g., the complete phrase {\small\tt severe mucositis} is annotated as an {\small\sf ADVERSE\_EVENT} but in case of {\small\tt Severe rhabdomyolysis}, only {\small\tt rhabdomyolysis} is annotated as an {\small\sf ADVERSE\_EVENT}. Hence, we carried out a lenient version of evaluation where a predicted {\small\sf ADVERSE\_EVENT} $AE_{predicted}$ is considered to be matching any gold-standard {\small\sf ADVERSE\_EVENT} $AE_{gold}$ if $AE_{predicted}$ contains $AE_{gold}$ as a prefix or suffix and $AE_{predicted}$ has at most one extra word as compared to $AE_{gold}$. E.g., even if $AE_{gold}$ = {\small\tt rhabdomyolysis} and $AE_{predicted}$ = {\small\tt Severe rhabdomyolysis}, we consider both of them to be matching. But if $AE_{gold}$ = {\small\tt severe mucositis} and $AE_{predicted}$ = {\small\tt mucositis}, we do not consider them to be matching because the predicted mention is missing a word which is expected as per the gold mention. This lenient evaluation leads to a much better performance as shown in Table~\ref{resultsADE}.

\section{Conclusion}
In this paper, we surveyed various techniques for jointly extracting entities and relations.
We first motivated the need for developing joint extraction techniques as opposed to traditional ``pipeline'' approaches. We then summarized more than a decade's work in joint extraction of entities and relations in the form of a table. In that table, we categorized techniques based on the approach they adopt for joint extraction, i.e. whether they employ joint inference or joint modelling or both. We further described some of the representative techniques for joint inference and joint modelling. We also described standard datasets and evaluation techniques; and summarized performance of the joint extraction approaches on these datasets. We presented a brief analysis of application of a general domain joint extraction approach on the ADE dataset from Biomedical domain. We believe that this survey would be useful for researchers as well as practitioners in the field of Information Extraction. Also, these joint extraction techniques would motivate new techniques even for other NLP tasks such as Semantic Role Labelling (SRL) where predicates and arguments can be extracted jointly.

\bibliographystyle{splncs04}
\bibliography{ref}
\end{document}